# Overlapping and Non-overlapping Camera Layouts for Robot Pose Estimation

**Mohammad Ehab Ragab**

**Informatics Department, Electronics Research Institute,**
**El-Dokki, Giza,12622, Egypt**

**Abstract**

We study the use of overlapping and non-overlapping camera layouts in estimating the ego-motion of a moving robot. To estimate the location and orientation of the robot, we investigate using four cameras as non-overlapping individuals, and as two stereo pairs. The pros and cons of the two approaches are elucidated. The cameras work independently and can have larger field of view in the non-overlapping layout. However, a scale factor ambiguity should be dealt with. On the other hand, stereo systems provide more accuracy but require establishing feature correspondence with more computational demand. For both approaches, the extended Kalman filter is used as a real-time recursive estimator. The approaches studied are verified with synthetic and real experiments alike.

## 1. Introduction

Real-time pose estimation is a popular research problem which aims at finding the location and orientation of moving objects or cameras synchronized with the captured frame. The applications are numerous ranging from intelligent vehicle guiding [1] to activity recognition [2], and behavior understanding [3].To get the pose or equivalently the ego-motion of a set of cameras on a robot in real time, we have to use recursive techniques. Two of the main recursive techniques are the particle filter, and the extended Kalman filter (EKF). Although the former is more advantageous in dense visual clutter, it requires increasing the computational cost to improve the performance [4]. However, the EKF has a satisfactory performance in indoor scenes.

The use of the EKF has two main aspects. The first is the number and arrangement of cameras. The second is the number and usage of filters. For example, a single camera and one EKF for both the pose and the structure are used in [5], [6], and in [7]. While four cameras arranged in two back-to-back stereo pairs, and one EKF for pose are used in [8]. One EKF is used for the 3D structure and the pose in [5], [6], [7], and [9]. A filter is used for the pose while a separate filter is used for each 3D point structure in [10] and [11]. In this way, the pose and the structure are decoupled.

In the field of computer vision, there are many challenges encountered in the robot pose estimation problem. For example, the availability and the reliability of the tracked features. Additionally, the robot motion brings about more difficulties such as the occlusion and reappearance of such features. Moreover, there are well known ambiguities such as the scale factor ambiguity related with using single cameras as well be shown below. The presence of such hardships justifies the use of printable 2-D circular marks as fiducials for tracking a mobile industrial robot in [12].

In this work, we use ordinary webcams to estimate the pose of a mobile robot within an unknown indoor scene. Particularly, we extend the work done in [13] and [14]. The motivation behind this is answering the following question: would the larger field of view of a non-overlapping layout compensate for the accuracy of stereo systems? For the non-overlapping layout, we use four cameras arranged individually on the platform of a moving robot. The axes passing through the camera centers of each back-to-back pair are perpendicular. This arrangement aims to maximize the joint field of view of all cameras. Each camera has multiple EKFs for the structure of tracked feature points and one EKF for the pose. For the overlapping layout, we use four cameras forming two stereo pairs put back-to-back on the robot as well. One EKF is used for the pose estimation. The 3D structure of the features fed to the filter is calculated by stereo triangulation based on the obtained pose which guarantees the consistency between the pose and the structure. For both layouts, the inputs to the system are the simultaneous frames taken by the calibrated cameras. The output is the real-time pose along the motion sequence of the robot. To avoid the effect of occlusion, we allow using a changeable set of features in each frame. The suggested camera layouts are shown in Fig. 1. The main contribution of this work is comparing four cameras arranged perpendicularly in a non-overlapping layout to the same number of cameras arranged as two back-to-back stereo pairs. Additionally, we propose a linear least squares method to solve for the scale factor ambiguity related to the non-overlapping single cameras. The rest of this paper is arranged as follows: the mathematical derivations and algorithms are explained in section 2, the simulations and real experiments are described in section 3, and the paper is concluded in section 4.

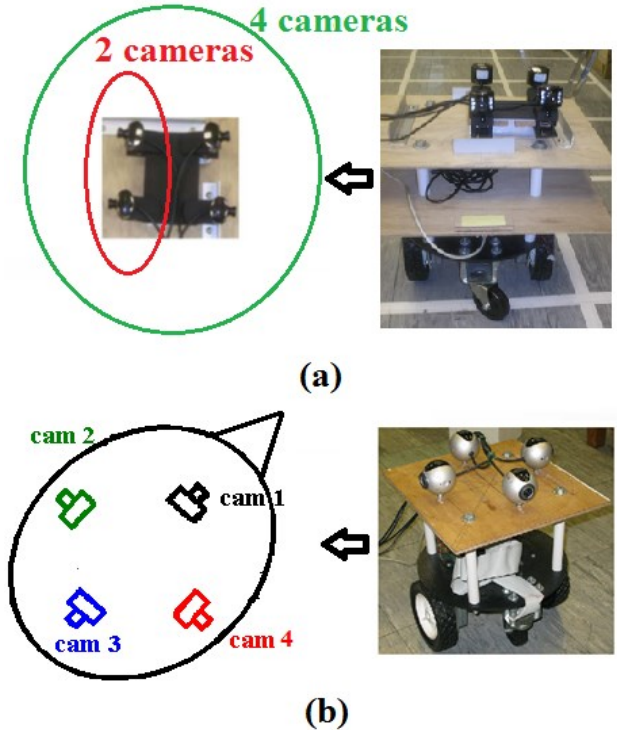

Fig.1 The two camera layouts studied: (a) the overlapping layout, and (b) the non-overlapping layout. To the right are nearly frontal views of the robot and cameras used for each layout. To the left are top-down schematics defining the compared approaches.

## 2. Method

2. 1 Mathematical Derivation

As shown in Fig. 2, initially before starting the motion, the center of the reference camera (camera 1 (cam 1)) is located at the origin of and aligned to the coordinate system ($x1; y1; z1$) with no rotation (for both layouts). Let the world coordinate system coincides with ($x1; y1; z1$, with respect to which the pose parameters are calculated). Let $D_k$ be a 3×1 displacement vector of cam k from cam 1 (with respect to the world coordinate system). Similarly, let $R_k$ be a 3×3 rotation matrix of cam k (with respect to the world coordinate system: $x1; y1; z1$). During the motion, at any general frame, $j$, the pose we want to estimate is given by the translation of cam 1, $d_j$, and its rotation matrix $R_j$ (with respect to the world coordinate system).

For the reference camera (cam 1), the camera coordinates, a 3 × 1 vector $P_{ij}$, of any 3-D feature $M_i$ (3 × 1 vector), at frame $j$ is given by:

$$P_{ij} = R_j^T (M_i - d_j) \qquad (1)$$

Where $T$ is the matrix transpose superscript.

Any other camera, the $k^{th}$ camera, has the following $j^{th}$ frame camera coordinates of $M_i$:

$$P_{ijk} = R_k^T R_j^T (M_i - d_j - R_j D_k) \qquad (2)$$

As mentioned above, initially cam 1 has $R_1$ and $D_1$ as the identity matrix and the zero vector respectively.

For the non-overlapping case, initially let cam k coincides with the coordinate system ($xk; yk; zk$, Fig. 2 (a)). Then, at frame $j$, let cam k be translated by the vector $l_{kj}$ and rotated by the rotation matrix $r_{kj}$ (with respect to the coordinate system $xk; yk; zk$). Due to the rigidity constraint, the rotation matrix $r_{kj}$ has an equivalent rotation $R_{kj}$ (brought back as occurring around $x1; y1; z1$). $R_{kj}$ is given by the change of basis [15]:

$$R_{kj} = R_k r_{kj} R_k^T \qquad (3)$$

If the pose of cam k is estimated ideally, then $R_{kj} = R_j$, however we take the median of all cameras to have a better estimate. We decompose the rotation matrix into the three rotation angles ($\alpha, \beta,$ and $\gamma$) around ($x1; y1; z1$) respectively and then take the median for each angle individually.

The situation is rather different for the location of cam k at frame $j$ (due to the scale factor ambiguity of single cameras). This location can be obtained in two ways shown in the following equation:

$$R_j D_k + S_j d_j = S_{kj} R_k l_{kj} + D_k \qquad (4)$$

The left hand side of equation (4) gives the displacement of cam k as shown in Fig. 2 (b) with adding the scale factor $S_j$ related to cam 1. The right hand side of this equation gives the displacement of cam k as shown in Fig. 2 (a) referred to the coordinate system ($xk; yk; zk$), scaled with the scale factor $S_{kj}$ related to cam k, and then having the axes transferred to that of ($x1; y1; z1$). To obtain these scale factors, we rearrange equation (4) to appear in the following form:

$$S_j d_j - S_{kj} R_k l_{kj} = (I - R_j) D_k \qquad (5)$$

Where $I$ is a 3×3 identity matrix. Equation (5) has three components in the directions of ($x1; y1; z1$). Additionally, there are three versions of this equation (for cam 2, cam 3, and cam 4). All components and versions can be assembled in the following system of linear equations:

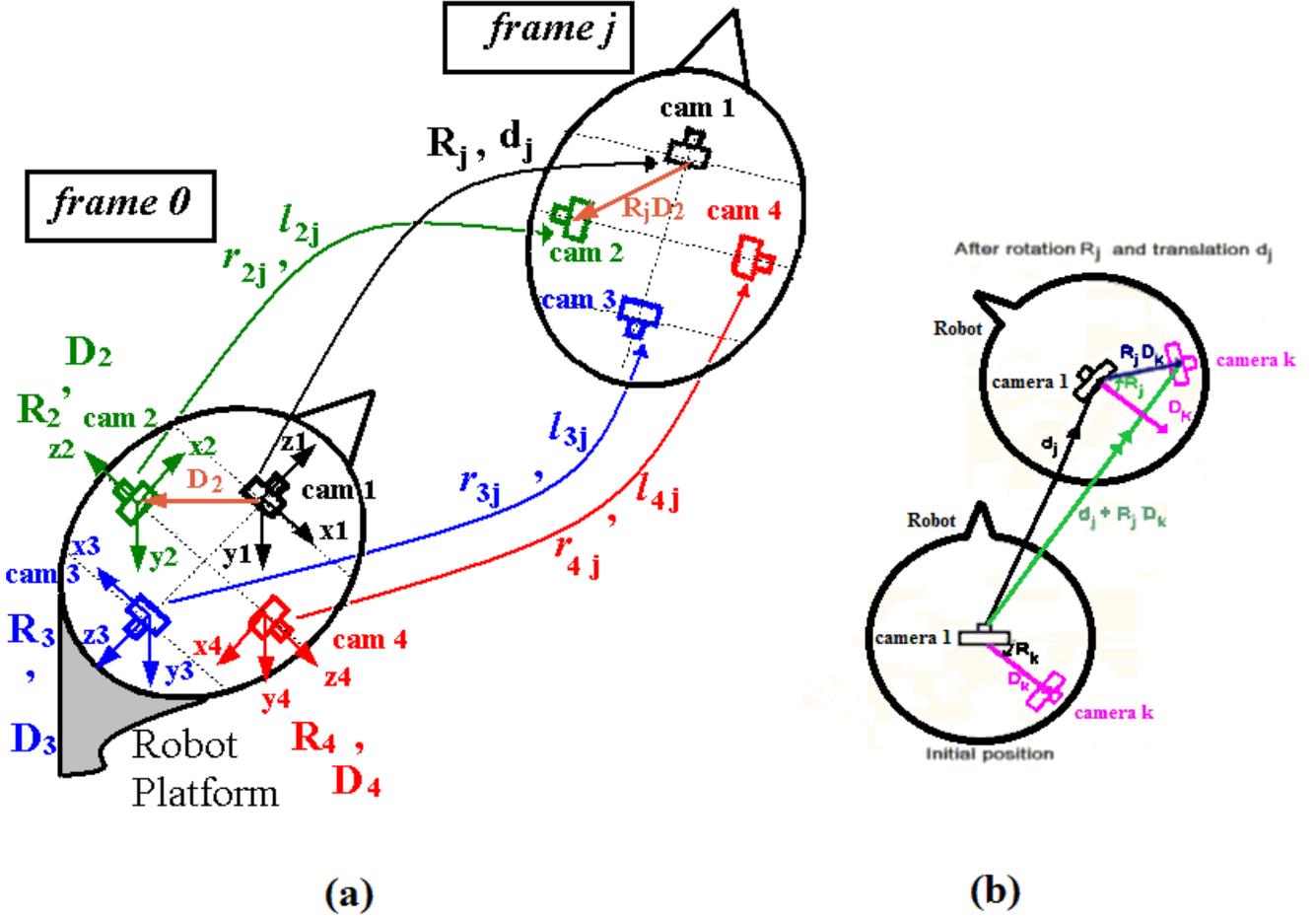

Fig. 2 Top-down view: effect of rotation and translation on the camera poses at any general frame, *j* (referred to the initial position). (a) Relative poses of non-overlapping layout. (b) Relative pose of any camera k (camera 2, camera3, or camera 4) referred to camera 1 (for both layouts). More details are mentioned in the text.

$$As = b \quad (6)$$

Where $A$ is a 9×4 matrix, $b$ is a 9×1 vector, and $s = [S_j\ S_{2j}\ S_{3j}\ S_{4j}]^T$ is a 4×1 vector of the unknown scales. The first row of $A$ is $[d_j|_{x1}\ R_2 l_{2j}|_{x1}\ 0\ 0]$ where $|_{x1}$ indicates the $(x1)$ component. The first element of vector $b$ is $(I - R_j)D_2|_{x1}$. The rest of the system of linear equations can be constructed in a similar way. The standard solution of such system in the least squares manner is given by:

$$s = (A^T A)^{-1} A^T b \quad (7)$$

Where $(\ )^{-1}$ indicates the matrix inversion.

In addition to the equations derived above to relate the multiple camera poses throughout the robot motion, the pose EKF is our estimator. It has a state space vector consisting of the six parameters of the pose, three translations in the directions of coordinate axes and three angles of rotations around them, and their derivatives (totally 12). A plant equation relates the current state, the previous one, and the plant noise. A measurement equation relates the measurements, features in pixels, to the measurement noise and the state measurement relation obtained using equations such as (1), and (2) above. A Jacobian is calculated for the EKF update step. This enables the filter to cope with the nonlinear model of the perspective camera. Besides using a pose EKF for each layout (overlapping and non-overlapping), an EKF is used to enhance the structure estimation of each 3D feature in the latter. The reason for this is that 3D triangulation for the structure calculation is not accurate enough for single cameras unlike the case of stereo pairs. More details about the EKF implementations can be found in [5], [6], [7], [16], and [17].

## 2.2. Algorithm

### 2.2.1 Overlapping (Stereo) Layout

Since the cameras remain rigidly fixed to the robot, the fundamental matrix ($F$) remains constant for each stereo-pair throughout the motion. Initially features are matched between a stereo-pair and tracked from frame to frame of the same camera provided that they verify $F$ of the stereo-pair (being close to the epipolar line, or rejected as outliers). The main steps of the algorithm are:
(1) For each stereo pair, find feature matches in the first frame and triangulate them to obtain the 3D structure.
(2) For each individual camera, track the features into the second frame, and obtain their pose using Lowe's method [18]. This step helps in starting the pose EKF as accurately as possible.
(3) Set the pose EKF to work feeding the feature locations in the 2D images as measurements, their structure, initial state space vector and covariance. The output is the current state vector (required pose and derivatives), and the current state covariance.
(4) Repeat step (3) recursively to the end of the sequence.
(5) As the robot moves, some features are occluded or come out of sight. When their number becomes less than a certain threshold (e.g. 50 features), backtrack to the previous frame, go to step (1) of this algorithm, then to step (3) (there is no need for step (2) since the previous state space vector is already available).

### 2.2.2 Non-overlapping Layout

(1) For each camera, find suitable features to track using a corner detector such as the KLT [19]. Initialize their 3D orthographically as lying on a plane whose depth is consistent with the environment around the robot. Pass each feature to a structure EKF to improve the estimation of its 3D structure.
(2) For each camera, track the features into the second frame, and obtain their pose using Lowe's method [18]. Pass each feature to a structure EKF.
(3) Set the pose EKF to work feeding the feature locations in 2D images as measurements, their structure, initial state space vector and covariance. The output is the current state vector, and covariance. Pass each feature to a structure EKF. Make use of the rigidity constraints of the multiple cameras to enhance the pose estimation (using the median of the estimated rotation angles and equation (7) for obtaining the scale factors).
(4) Repeat step (3) recursively to the end of the sequence.
(5) When the number of features becomes less than a certain threshold (e.g. 50 features), backtrack to the previous frame, go to step (1) of this algorithm, then to step (3) (there is no need for step (2) since the previous state space vector is already available).

## 3. Experiments

### 3.1. Simulations

We put four non-overlapping cameras as shown in Fig. 1 (b) on a robot. Then, we pair the camera in the front and the camera in the back with another camera to form two stereo pair as in Fig. 1 (a). Therefore, in all we have six cameras. The baselines for stereo cameras are 0.1 meter which is as the same as the distance from cam 1 to any other camera. Each camera has a 6 mm focal length, and 640×480 resolution. The robot moves with six degrees of freedom: translations ($t_x, t_y,$ and $t_z$), and rotation angles ($\alpha, \beta,$ and $\gamma$) with respect to the coordinate system attached to the reference camera (cam 1). The translations are taken randomly from ±0:005 to ±0:015 meter, and the rotation angles are taken randomly from ±0:005 to ±0:02 radian. A zero-mean Gaussian noise with 0.5 standard deviation is added to the measurements fed to the EKFs. The 3D features (whose number is 10,000) are distributed randomly to form a spherical shell extending from 0.667 to one meter. The simulations are run 1,500 times with a sequence of 100 frames captured by each camera. We have calculated the absolute errors for the six pose parameters using both overlapping and non-overlapping layouts. For the overlapping case, as shown in Fig. 1 (a), we consider '2 cameras' formed by the front stereo pair, and '4 cameras' formed by the front and back stereo pairs. For the non-overlapping case, as shown in Fig. 1 (b), we consider 'cam 1', 'cam 2', 'cam 3', 'cam4', and 'RC' verifying the rigidity constraints among the multiple cameras (as explained in section 2 above). Table 1 shows the simulation results.

Table 1: Average absolute error of pose values per frame (simulations)

| Method | $t_x$ m | $t_y$ m | $t_z$ m | $\alpha$ rad | $\beta$ rad | $\gamma$ rad |
|---|---|---|---|---|---|---|
| 4 cameras | .0005 | .0005 | .0021 | .0005 | .0004 | .0015 |
| 2 cameras | .0055 | .0132 | .0042 | .0137 | .0055 | .0026 |
| cam 1 | .0252 | .0206 | .0337 | .0204 | .0260 | .0040 |
| cam 2 | .2973 | .3731 | .3828 | .0864 | .2724 | .2791 |
| cam 3 | .0238 | .0211 | .0390 | .0203 | .0249 | .0045 |
| cam 4 | .2846 | .3740 | .3770 | .0879 | .2750 | .2825 |
| RC | .0211 | .0694 | .0746 | .0288 | .0624 | .0553 |

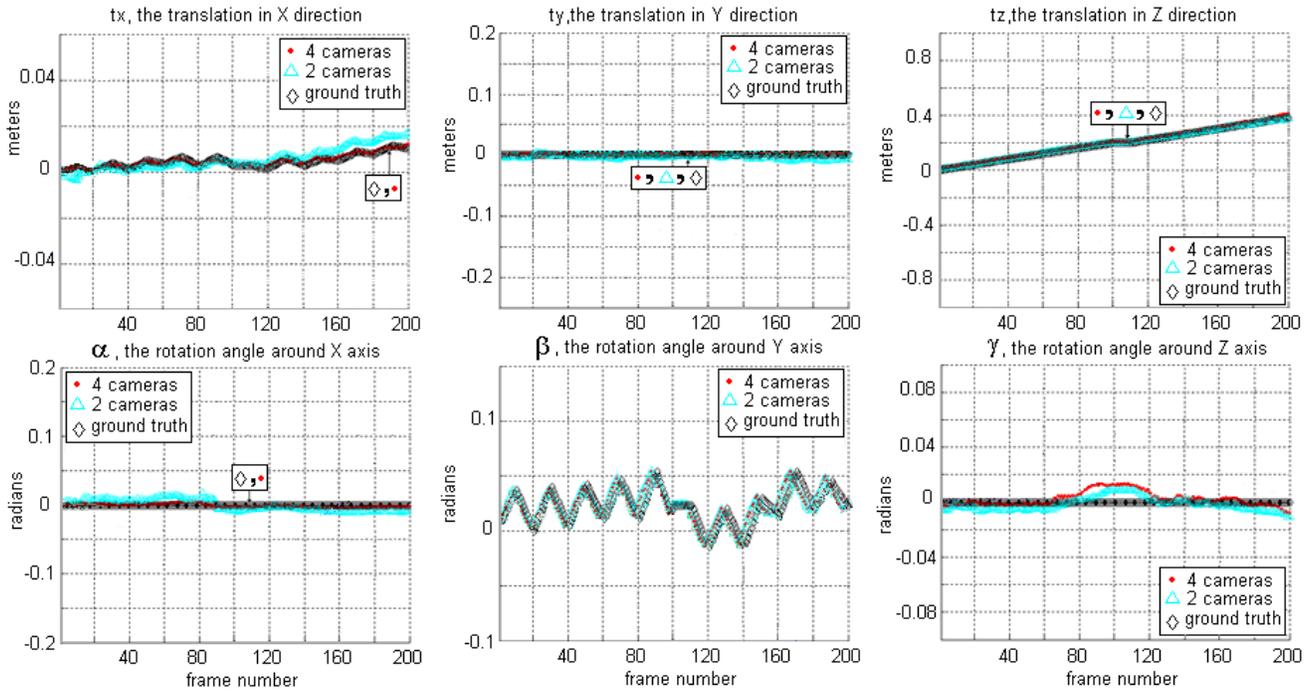

Fig. 3 Real experiments of pose estimation (overlapping layout), upper row: translation components, lower row: rotation angles. '2 cameras' indicates using the front stereo pair, '4 cameras' indicates using both the front and the back stereo pairs.

3.2. Real Experiments

Two sets of real experiments have been carried out using four webcams with resolution 320×240 each. The first set uses the overlapping layout (4 cameras, and 2 cameras). The second set uses the non-overlapping layout (cam 1, cam 2, cam 3, cam 4, and RC). A sequence of 200 frames has been taken simultaneously by each camera while the robot is following a motion of mixed rotation and translation. It deserves mentioning that carrying out a set of real experiments for each layout is necessary for a better synchronization of the frames captured by the multiple cameras. The experimental results are shown in Fig. 3, and Fig. 4. Samples of the captured sequences are shown in Fig. 5.

## 4. Discussion and Conclusions

The simulations, in Table 1, show that the overlapping layout is more accurate than the non-overlapping one in estimating the robot pose parameters. The reason for this is that the obtained pose is affected by the 3D structure of features fed into the pose EKF. This structure is obtained more accurately for the overlapping layout based on stereo triangulation. On the other hand for single cameras, the depth of the features is assumed orthographically initially. Although this assumption is recursively enhanced using the structure EKFs, its effect propagates (at least through the early frames of the sequence). Additionally using two stereo pairs, '4 cameras', enhances the accuracy further. This is expected due to having more stereo information which improves the obtained pose parameters.

For the non-overlapping layout, the reference camera, 'cam 1', and its back-to-back camera, 'cam 3', are the most accurate. This is logical since the reference camera is located exactly on the spot of robot platform whose pose is estimated. For 'cam 3' (its back-to-back camera), the translational and rotational pose parameters are not correlated [20]. This explains the accuracy of its pose parameters. In contrast, this is not the case for the two perpendicular cameras, 'cam 2' and 'cam 4'. Their locations do not guarantee decoupling the translational and rotational pose parameters. Making use of the rigidity constraints, 'RC', is more accurate than both perpendicular cameras. However, compared to 'cam 1' and 'cam 3', it is only more accurate for '$t_x$'. We expect that selecting good spots for 'cam 2' and 'cam 4' would enhance 'RC' a lot.

As shown in Fig. 3, and Fig. 4, the real experiments agree with the simulations. For most pose parameters, both '4 cameras' and '2 cameras' are close to the ground truth. In the non-overlapping case, 'cam 1' and 'cam 3' have better performances than the other two cameras (though not exactly following the ground truth). Imposing the rigidity constraints in 'RC' improves the performance for most pose parameters. Its validity is even more obvious in the real experiment where the initial assumed depths are not as accurate as in the simulations (for any individual camera).

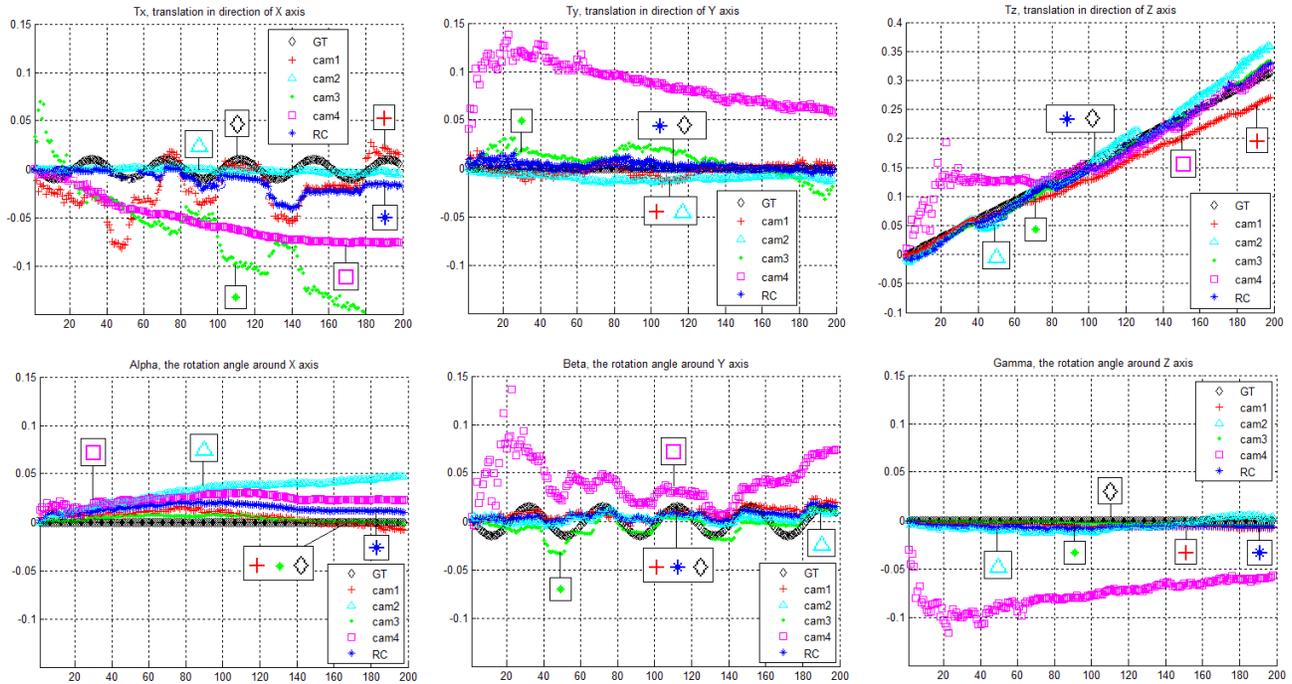

Fig. 4 Real experiments of pose estimation (non-overlapping layout), upper row: translation components, lower row: rotation angles. 'cam 1' indicates using camera 1 individually (which is the case for other single cameras), 'GT' indicates the ground truth, and 'RC' indicates using the multiple cameras to verify the rigidity constraints.

To sum up, the overlapping layout verifies more accurate pose estimation. However, this requires the stereo calibration of cameras and stereo matching of corresponding features initially (at the first frame), and whenever it is needed throughout the sequence. On the other hand, using non-overlapping layout provides accurate tracking of short base-line features. Besides avoiding the stereo matching, it increases the information captured by the multiple cameras since they cover a larger field of view than the stereo systems. Furthermore, the approach can be easily deployed for parallel processing. Moreover, there is a room for improvement within it by selecting good spots for 'cam 2' and 'cam 4' (instead of putting them perpendicularly) which would boost the accuracy of the approach imposing the rigidity constraints. This would be further emphasized by adopting a more efficient optimization approach than the straightforward least squares.

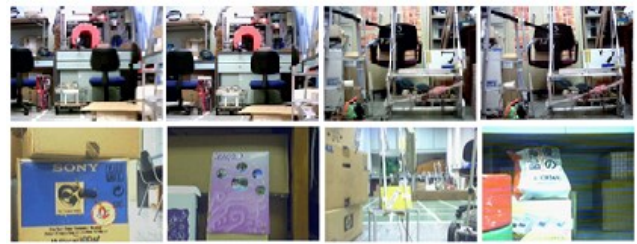

Fig. 5 Samples of the frames taken by the cameras. Upper row belongs to the overlapping layout (from left to right: front stereo pair, then back stereo pair). Lower row belongs to the non-overlapping layout (from left to right: cam 1, cam 2, cam 3, and cam 4).

**Mohammad Ehab Ragab** is a Researcher at the Informatics Department, the Electronics Research Institute in Egypt. He received the B.Sc., M.Sc., degrees from Ain Shams University, and Cairo University respectively. He obtained his Ph.D. from the Chinese University of Hong Kong (CUHK) in 2008. During his undergraduate studies, he received the "Simplified Scientific Writing Award" from the Academy of Scientific Research and Technology. He was awarded the "IDB Merit Scholarship" for conducting his Ph.D. studies. He taught courses in some distinguished Egyptian universities, and published in international journals and conferences. His research interests include: Computer vision, Robotics and Image Processing.